\documentclass[letterpaper, 10 pt, conference]{ieeeconf}  

\IEEEoverridecommandlockouts                              

\usepackage{colortbl}
\usepackage{xcolor}
\usepackage[utf8]{inputenc} 
\usepackage[T1]{fontenc}    
\usepackage{hyperref}       
\usepackage{url}            
\usepackage{booktabs}       
\usepackage{amsfonts}       
\usepackage{nicefrac}       
\usepackage{microtype}      
\usepackage{xcolor}         
\usepackage{graphicx}
\usepackage{multirow}
\usepackage{bm}
\usepackage{pifont}
\usepackage{subfigure}
\usepackage{amsmath}
\usepackage[ruled]{algorithm2e}
\usepackage{multirow}
\usepackage{subfiles}
\usepackage{hyperref}       
\usepackage{overpic}
\usepackage{mathrsfs}
\usepackage{geometry}
\title{\LARGE \bf
Semi-Supervised Learning for Visual Bird's Eye View Semantic Segmentation 
}

\author{Junyu Zhu$^{1*}$, Lina Liu$^{2*}$, Yu Tang$^{3}$, Feng Wen$^{3}$, Wanlong Li$^{3\dag}$ and Yong Liu$^{1\dag}$ 
\thanks{$^{1}$Junyu Zhu and Yong Liu are with the Institute of Cyber-Systems and Control, Zhejiang University, Hangzhou, China. E-mail:junyuzhu@zju.edu.cn, yongliu@iipc.zju.edu.cn.}
\thanks{$^{2}$Lina Liu is with China Mobile Research Institute, Beijing, China. E-mail: liulina0601@gmail.com.}
\thanks{$^{3}$Wanlong Li, Yu Tang, and Feng Wen are with Noah’s Ark Lab, Huawei Technologies, Beijing, China. E-mail:\{liwanlong, tangyu17, wenfeng3\}@huawei.com.}
\thanks{$^{*}$Equal contribution.$^{\dag}$Corresponding author.}
}

\begin{document}

\maketitle
\thispagestyle{empty}
\pagestyle{empty}

\begin{abstract}
Visual bird's eye view (BEV) semantic segmentation helps autonomous vehicles understand the surrounding environment only from front-view (FV) images, including static elements (e.g., roads) and dynamic elements (e.g., vehicles, pedestrians). However, the high cost of annotation procedures of full-supervised methods limits the capability of the visual BEV semantic segmentation, which usually needs HD maps, 3D object bounding boxes, and camera extrinsic matrixes. In this paper, we present a novel semi-supervised framework for visual BEV semantic segmentation to boost performance by exploiting unlabeled images during the training. A consistency loss that makes full use of unlabeled data is then proposed to constrain the model on not only semantic prediction but also the BEV feature. Furthermore, we propose a novel and effective data augmentation method named conjoint rotation which reasonably augments the dataset while maintaining the geometric relationship between the FV images and the BEV semantic segmentation. Extensive experiments on the nuScenes dataset show that our semi-supervised framework can effectively improve prediction accuracy. To the best of our knowledge, this is the first work that explores improving visual BEV semantic segmentation performance using unlabeled data. The code is available at 
\href{https://github.com/Junyu-Z/Semi-BEVseg}{\textcolor{red}{https://github.com/Junyu-Z/Semi-BEVseg}}.

\end{abstract}

\section{Introduction}

Bird's eye view (BEV) semantic segmentation is a powerful representation of the surrounding environment, which can assist mobile robots such as autonomous vehicles in perceiving the surroundings of static road layouts and dynamic objects(e.g., vehicles, pedestrians). With rich information and absolute scales, BEV semantic segmentation can directly connect with many downstream tasks, such as path planning and motion control. Recently, vision-based methods~\cite{PON,gong2022gitnet,saha2022translating,zhou2022cross,lu2019monocular} that infer BEV semantic segmentation only from cameras have been developed to reduce the cost of sensors. 

A visual BEV semantic segmentation model generally consists of three components~\cite{li2022delving}: a backbone network as a visual feature extractor, a view transformer module for getting the BEV feature from the front-view (FV) feature, and a segmentation decoder to predict semantic segmentation from the BEV feature. And most of the existing BEV semantic segmentation methods are full-supervised, mainly focusing on exploring new view transform approaches~\cite{philion2020lift, PON,zhou2022matrixvt}, integrating temporal cues~\cite{saha2021enabling,li2022bevformer}, and designing more complex segmentation decoders~\cite{saha2022translating,gosala2022bird}. However, these methods rely heavily on the accessibility and quantity of labeled data that needs high costs for constructing HD maps, annotating 3D object bounding boxes, and capturing camera extrinsic parameters. Compared with annotation, collecting unlabeled images requires less labor. Therefore, in this work, we are motivated to study semi-supervised learning based BEV semantic segmentation from monocular images to boost the performance by exploiting unlabeled data. 

\begin{figure}[t]
    \centering
    \includegraphics[scale=0.32]{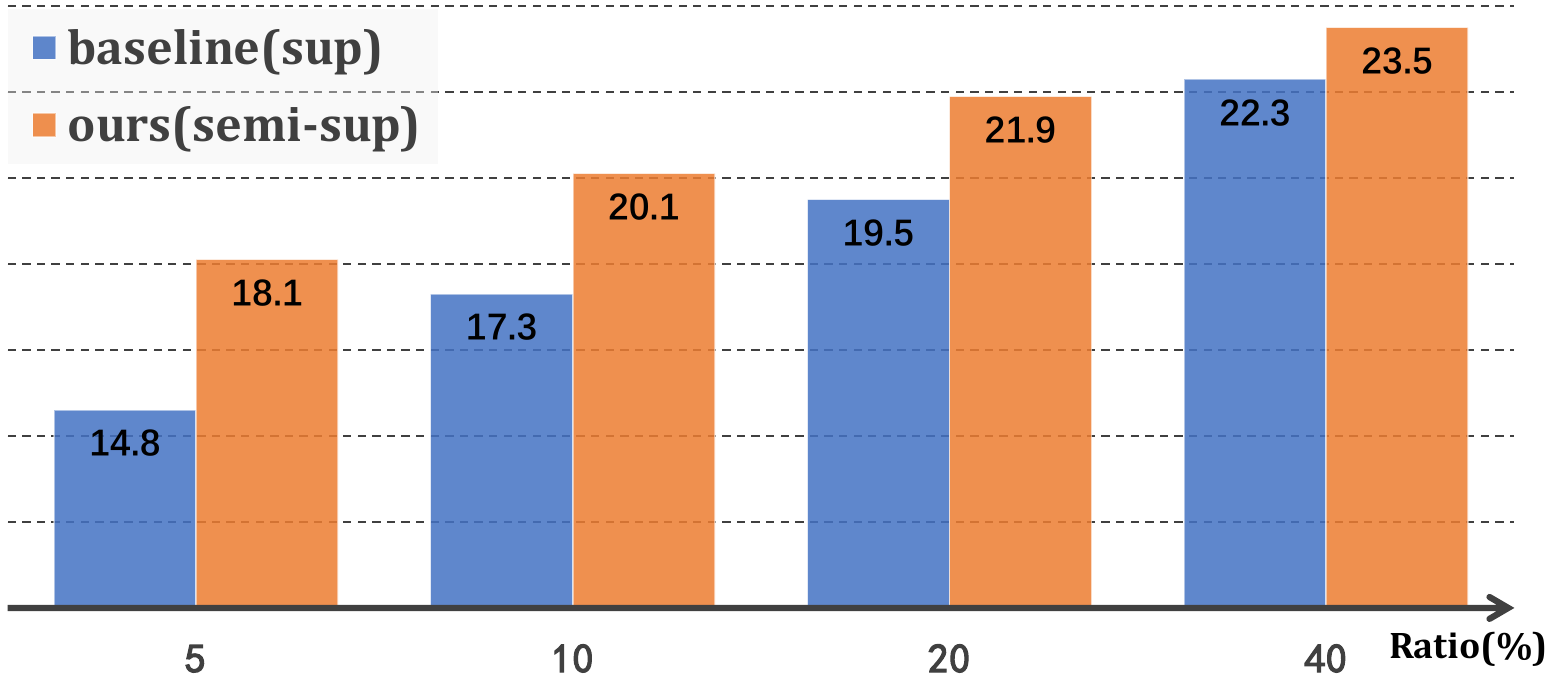}
    \vspace{-0.4cm}
    \caption{mIoU(\%) on the nuScenes dataset between our semi-supervised framework and supervised baseline using different label ratios.}\label{comparison}
    \vspace{-0.5cm}
\end{figure}

While many works have explored semi-supervised learning for conventional 2D semantic segmentation, semi-supervised visual BEV semantic segmentation is rather underexplored. Following the common consistency regularization in semi-supervised learning, we propose a consistency loss that restricts the model with perturbations on unlabeled images. Inspired by~\cite{huang2021bevdet}, in addition to semantic segmentation consistency, we use additional consistency of BEV feature for further improvement. And to excavate the spatial consistency of the BEV feature, we use horizontal flipping as the perturbation rather than color jitter which is typical for semi-supervised 2D semantic segmentation.

Apart from using the above consistency regularization on unlabeled data, we also explore improving the quantity and diversity of the dataset for better performance. Although several well-designed and effective data augmentation methods~\cite{ghiasi2021simple,xiaopolarmix} have been proposed for 2D/3D semantic segmentation, there is no relevant research in the visual BEV semantic segmentation field. Unlike pixel-aligned 2D/3D semantic segmentation, the complex geometric relationship of the projection between FV images and BEV semantic segmentation maps makes data augmentation harder. Through geometric intuition and mathematical analysis, we propose a novel data augmentation method called conjoint rotation for this task. And it benefits not only our semi-supervised framework but also the full-supervised model.

Following the conventions in semi-supervised tasks, we conduct experiments on nuScenes~\cite{caesar2020nuscenes} 
dataset with different ratios of labels and demonstrate that our semi-supervised framework can effectively improve performance by relatively \textgreater$10\%$ on average with the unlabeled data as shown in Fig.~\ref{comparison}. Moreover, extensive ablation studies are also conducted to prove the effectiveness of each component. We hope this work can be a stepping-stone for future research in this field.

To summarize, our main contributions are as follows:
\begin{itemize}
  \item We dig into visual BEV semantic segmentation with limited labels and offer the first semi-supervised BEV semantic segmentation framework that enhances the performance using unlabeled data.
  \item We propose a consistency loss exploiting unlabeled data to restrict the model on semantic segmentation and the BEV feature.
  \item We design a novel data augmentation method for visual BEV semantic segmentation, and it works well on our semi-supervised framework and full-supervised model.
  \item The proposed framework achieves relatively \textgreater$10\%$ average improvements over the full-supervised baseline on the nuScenes.
\end{itemize}

\section{Related works}

\subsection{Visual BEV Semantic Segmentation}
Visual BEV semantic segmentation is a task of using FV images to predict BEV semantic segmentation. Via homography transformations, \cite{sengupta2012automatic,bev_on_board} use inverse perspective mapping (IPM) to map FV images/features onto the BEV plane. This approach relies heavily on the plane hypothesis, so it easily fails for objects that lie above the BEV plane, such as cars and pedestrians. VED~\cite{lu2019monocular} uses the fully-connected bottleneck layer to realize the feature transformation from the front view to the BEV. Due to the lack of available ground truth data, early methods rely on various weak supervision. And with the emergence of the nuScenes dataset~\cite{caesar2020nuscenes} that contains HD maps, 3D object bounding boxes, and much image data from six calibrated cameras in different scenes, visual BEV semantic segmentation develops rapidly. Based on view transformation (VT) strategies, different methods can be divided into the following categories:




MLP-based VT~\cite{PON, saha2022translating, gong2022gitnet} is based on the geometric correspondence between the vertical lines in the image and polar rays in BEV.
2D-to-3D-based VT~\cite{dwivedi2021bird, philion2020lift} gets BEV feature by explicit or implicit depth estimation.
3D-to-2D-based VT~\cite{roddick2018orthographic,li2022bevformer,liuvision}  projects 3D points from the BEV plane onto the 2D image plane to get corresponding features.
Transformer-based VT~\cite{zhou2022cross,bartoccionilara,peng2023bevsegformer,li2022bevformer} is another ready solution for transforming features from the front view to the BEV by implicit geometric reasoning.

Although impressive results have been achieved by recent fully-supervised methods, requiring time-consuming and laborious labeling is a common shortcoming. Gao et al.~\cite{gao2022s2g2} present a framework that can be trained with both labeled and unlabeled data but fails to improve performance with unlabeled data. And their work focuses on estimating road layout but no dynamic elements. In this work, under a more challenging setting, we dig into underexplored semi-supervised learning in visual BEV semantic segmentation to improve performance by exploiting unlabeled data.

\begin{figure*}[t]
    \centering
    \includegraphics[scale=0.48]{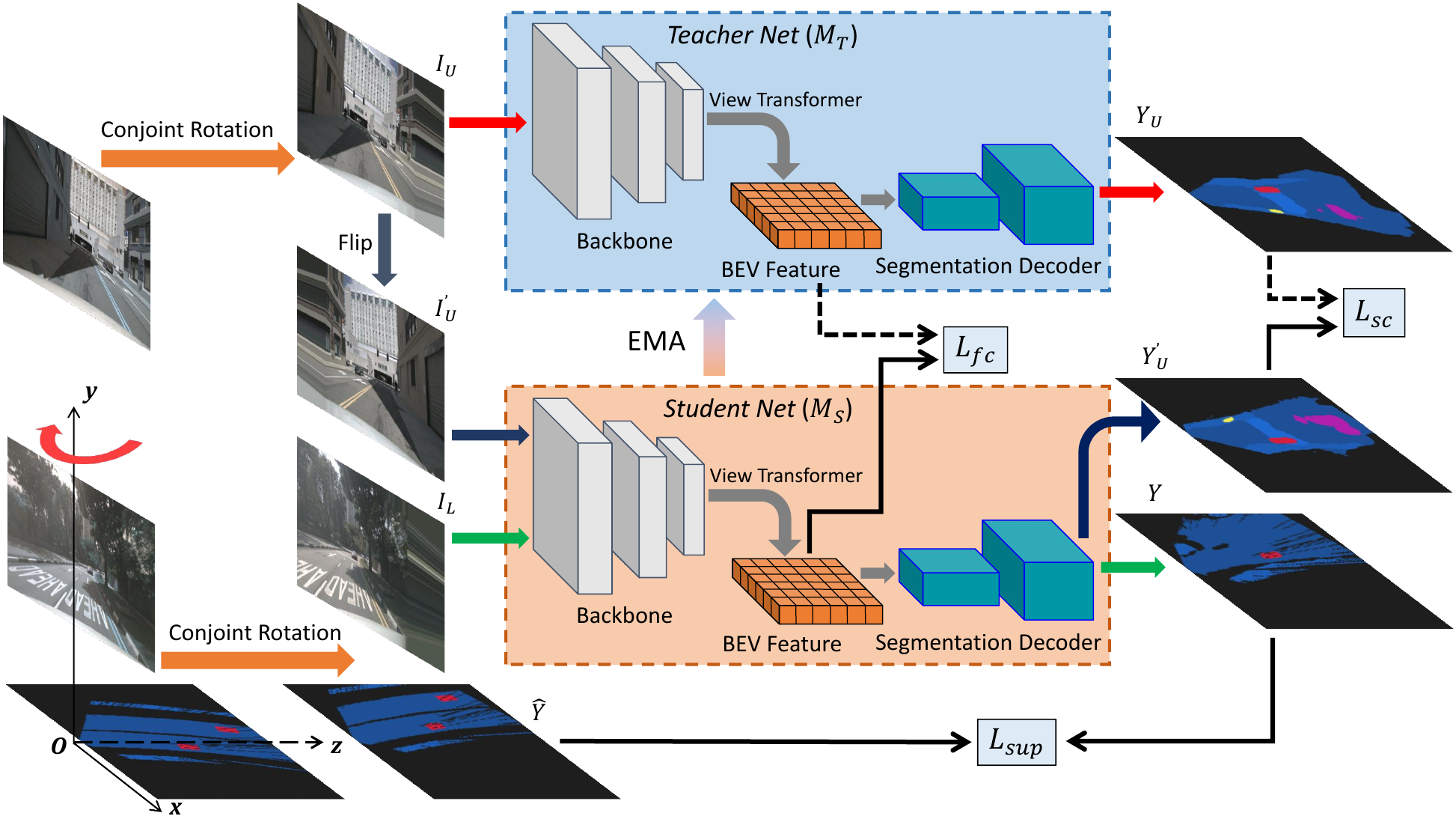}
    \vspace{-0.3cm}
    \caption{Framework overview. By our proposed conjoint rotation, the labeled and unlabeled data are first augmented to get $I_{L}$, $\hat{Y}$, and $I_{U}$. Immediately after that, $Y$ and $Y_{U}$ are predicted by the Student Net $M_{S}$ and Teacher Net $M_{T}$, respectively. Meanwhile, $M_{S}$ predicted $Y^{'}_{U}$ from flipped image $I^{'}_{U}$. Note that the view transformer of $M_{S}$ and $M_{T}$ needs the camera intrinsic matrix $K$ as input, and $K$ would also be changed when flipping the image. The feature consistency loss $L_{fc}$ is computed from the L2 loss of BEV features of $I_{U}$ and $I^{'}_{U}$. And the segmentation consistency loss $L_{sc}$ is computed from the L2 loss of BEV semantic segmentation, $Y_{U}$ and $Y^{'}_{U}$. Also, the supervised loss $L_{sup}$ is computed between $Y$ and $\hat{Y}$. After $M_{S}$ is updated with gradient descent using the above losses, $M_{T}$ is updated as an exponential moving average (EMA) of $M_{S}$. The Teacher Net can perform better than the Student Net after the training with proper hyper-parameters.}\label{pipline}
    \vspace{-0.5cm}
\end{figure*}

\subsection{Semi-Supervised 2D Semantic Segmentation}
Inspired by the progress of semi-supervised learning in the image classification field, semi-supervised semantic segmentation for the 2D image has been explored by many works in these years. To force the decision boundary to lie in the low-density area, many works~\cite{french2021colour,olsson2021classmix,french2019semi,verma2022interpolation} utilize a common strategy, consistency regularization. Pseudo-labeling~\cite{kalluri2019universal,chen2021semi,ouali2020semi} is another effective technique. 

In this work, we apply consistency regularization to the semi-supervised visual BEV semantic segmentation task and propose a consistency loss that acts on both semantic segmentation and the BEV feature.

\subsection{Data Augmentation}
Data augmentation is a practical approach to improving generalization ability and has been explored in many fields, including image classification~\cite{zhong2020random,yun2019cutmix}, 3D point cloud semantic segmentation~\cite{xiaopolarmix}, and 2D semantic segmentation~\cite{ghiasi2021simple}. In the visual BEV semantic segmentation field, there is no relevant work currently, to the best of our knowledge.

In this work, out of geometric intuition and mathematical analysis, we propose a new data augmentation named conjoint rotation that is effective for this task.

\section{Method}


For the visual BEV semantic segmentation task, we need to predict a semantic segmentation map $Y$ from the given FV image $I$ with its corresponding camera intrinsic matrix $K$. In this paper, each pixel of $Y\in p^{C\times Z\times X}$ describes the probability of $C$ categories, such as drivable area, walkway, pedestrian, and car. And different types in a BEV semantic map may appear in the same pixel, which is different from the setting of some existing works~\cite{philion2020lift,li2022bevformer,zhou2022cross}. Under the semi-supervised setting, the training set consists of a labeled set $D_{L}=\{(I^{0}_{L}, K^{0}_{L}, \hat{Y}^{0}), (I^{1}_{L}, K^{1}_{L}, \hat{Y}^{1}), ..., (I^{i}_{L}, K^{i}_{L}, \hat{Y}^{i}), ...\}$ and an unlabeled set $D_{U}=\{(I^{0}_{U}, K^{0}_{U}), (I^{1}_{U}, K^{1}_{U}), ..., (I^{i}_{U}, K^{i}_{U}), ...\}$. And we aim to exploit $D_{L}\cup D_{U}$ to train a model that performs better than only trained on $D_{L}$. An overview of the proposed framework is illustrated in Fig.~\ref{pipline}.

In this work, we follow the Mean Teacher~\cite{tarvainen2017mean} that is originally proposed for image classification and extend it to the more taxing task of visual BEV semantic segmentation. We design a segmentation consistency loss $L_{sc}$ and a feature consistency loss $L_{fc}$ for consistency regularization. Furthermore, we propose a novel data augmentation method called conjoint rotation to improve performance. 

In the following subsections, we first give a brief introduction to the visual BEV semantic segmentation model in Sec.~\ref{model}. In Secs.~\ref{supervised_loss},\ref{segmentation_consistency},\ref{feature_consistency}, we successively describe the supervised loss $L_{sup}$, segmentation consistency loss $L_{sc}$, and BEV feature consistency loss $L_{fc}$. And we present our proposed conjoint rotation in Sec.~\ref{conjoint_rotation}. Finally, Sec.~\ref{training_process} summarizes the training process.

\subsection{Visual BEV semantic segmentation Model}\label{model}
Generally, a visual BEV semantic segmentation model $M$ first uses a backbone network to extract the FV feature from the given FV image $I\in R^{3\times H\times W}$. The model gets the BEV feature from the FV feature through a view transformer that is usually related to the camera intrinsic matrix $K\in R^{3\times 3}$. Finally, using a segmentation decoder, the model predicts BEV semantic segmentation $Y\in p^{C\times Z\times X}$ from the BEV feature. In this work, our framework uses two models with the same structure called Teacher Net $M_{T}$ and Student Net $M_{S}$. Their parameters are separately randomly initialized except the pretrained backbone network, and the $M_{T}$ performs better after the training process.

\subsection{Supervised Loss}\label{supervised_loss}
Following state-of-the-art methods~\cite{saha2021enabling,saha2022translating,gong2022gitnet}, we use the same Dice loss as the supervised loss $L_{sup}$ for labeled data. The $L_{sup}$ is defined over $C$ classes and $N$ pixels:
\begin{equation}
    L_{sup}=1-\frac{1}{\vert C\vert}\sum_{k=1}^{C}\frac{2\sum_{i}^{N}\hat{y}_{i}^{k}y_{i}^{k}}{\sum_{i}^{N}\hat{y}_{i}^{k}+y_{i}^{k}+\epsilon}.
\end{equation}
where $\hat{y}_{i}^{k}$ is the target binary variable grid cell of $\hat{Y}$, $y_{i}^{k}$ is the predicted probability variable of $Y$, and $\epsilon$ is set as 1e-5 to prevent the denominator from being zero.

\subsection{Segmentation Consistency Loss}\label{segmentation_consistency}
For the unlabeled data, we calculate the segmentation consistency loss $L_{sc}$ between semantic segmentations, $Y_{U}$ and $Y^{'}_{U}$, from the unlabeled data $\{I_{U}, K_{U}\}$ and the horizontally flipped version $\{I^{'}_{U}, K^{'}_{U}\}$. The $Y_{U}$ and $Y^{'}_{U}$ are the output class probability matrixes after the last sigmoid function. Because of the geometric relationship between the FV and the BEV, the consistency of BEV semantic segmentation of the original image and the flipped one is natural. Using L2 distance $\Vert\cdot\Vert_{2}$ and horizontally flipping operation $\Phi$, the $L_{sc}$ is formulated as:
\begin{equation}
    L_{sc}=\Vert \widetilde{Y}_{U}-\Phi(Y^{'}_{U})\Vert_{2}.
\end{equation}
where $\widetilde{Y}_{U}$ means that the gradient of $Y_{U}$ is detached.

\subsection{BEV Feature Consistency Loss}\label{feature_consistency}
In BEVDet~\cite{huang2021bevdet}, Huang et al. conducted common 2D augmentation operations, including random flipping, scaling and rotating on both the BEV feature and the 3D object detection targets for boosting the detection performance. Their augmentation strategy actually indicates spatial correspondence between the BEV feature and BEV position. And we further find the spatial correspondence between the BEV feature and BEV semantic segmentation can also be established. In other words, when two semantic segmentation maps are symmetric, their BEV features should also be symmetric. Thus apart from applying consistency in BEV semantic segmentation, we design a feature consistency loss $L_{fc}$ for the BEV feature to refine the consistency:
\begin{equation}
    L_{fc}=\Vert \widetilde{F}_{U}-\Phi(F^{'}_{U})\Vert_{2}.
\end{equation}
where $F_{U}$ and $F^{'}_{U}$ are the BEV features of $I_{U}$ and $I^{'}_{U}$. $\Vert\cdot\Vert_{2}$ is the L2 distance. $\Phi$ denotes the horizontally flipping operation and $\widetilde{F}_{U}$ means that the gradient of $F_{U}$ is detached.

\subsection{Conjoint Rotation for Data Augmentation}\label{conjoint_rotation}
Data augmentation can effectively improve the quantity and diversity of the training set to boost performance. There are no specially designed data augmentation methods on the visual BEV semantic segmentation, and only some simple methods, e.g., horizontal flipping and color jitter are used, to the best of our knowledge. it's mainly because the geometric relationship between the FV image and the BEV semantic segmentation is more complex than in the pixel-aligned 2D/3D tasks. The existing methods can easily destroy the spatial position relationship of corresponding pixels, making the view transformer more challenging to be trained. 

\begin{figure}[h]
    \centering
    \vspace{-0.2cm}
    \includegraphics[scale=0.25]{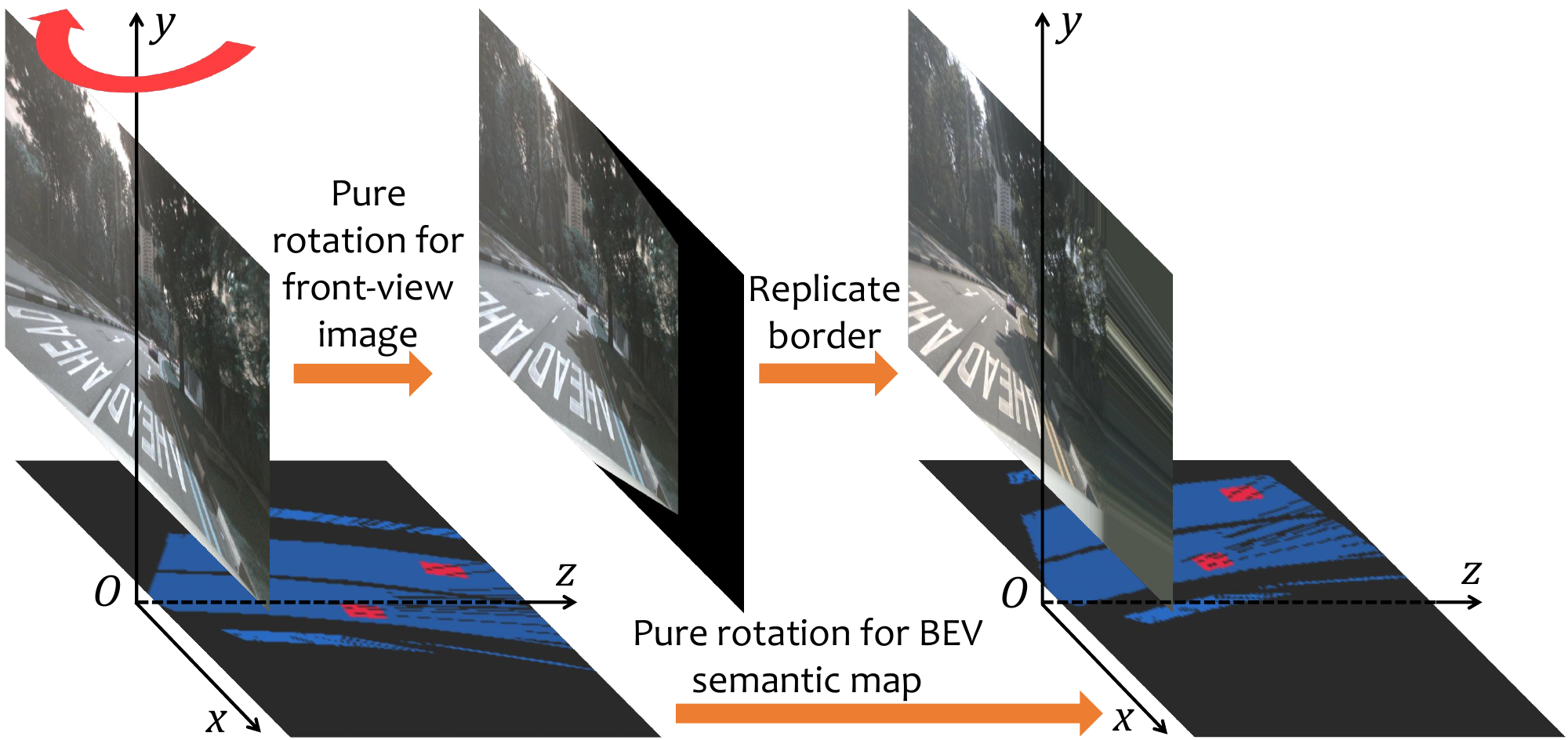}
    \vspace{-0.2cm}
    \caption{Illustration of conjoint rotation.}\label{conjoint_rotation_process}
    \vspace{-0.3cm}
\end{figure}

We find that conjointly rotating the FV image and the GT BEV semantic segmentation map can reasonably augment the dataset without damaging the geometric relationship. As shown in Fig.~\ref{conjoint_rotation_process}, with the random angle $\alpha$ sampled from a pre-determined interval $\left[-\alpha_{max},\alpha_{max}\right]$, we rotate the GT BEV semantic segmentation map and the FV image along a y-axis that is vertical to the BEV plane and passes through the origin of camera coordinate system. 

The above rotation is a pure rotation for the camera, so the rotated FV image can be obtained using a homography transformation $H_{1}$ that merely relates to the camera intrinsic matrix $K$ and angle $\alpha$ by forward warping operation. And according to~\cite{bian2021auto}, the transformation brought by $H_{1}$ can be expressed using the follow equations:
\begin{equation}\label{homography_matrix}
    \left\{
    \begin{array}{l}
        u_{2} = (h_{11}u_{1}+h_{12}v_{1}+h_{13})/(h_{31}u_{1}+h_{32}v_{1}+h_{33}) \\
        v_{2} = (h_{21}u_{1}+h_{22}v_{1}+h_{23})/(h_{31}u_{1}+h_{32}v_{1}+h_{33})
    \end{array},
    \right.
\end{equation}
where the $(u_{1},v_{1})$ and $(u_{2},v_{2})$ respectively denotes the pixel coordinate in the original and transformed images, and $h_{ij}$ is only determined by $K$ and $\alpha$. The above forward warping operation will introduce black edges in the transformed image. Such black edges can lower the improvement of the conjoint rotation, and we find replicating the border after the homography transformation can work better, as shown in our experiments in Sec.~\ref{conjoint_rotation_ablation}.

Furthermore, the perpendicular relationship between the BEV plane and the y-axis makes the rotation of the BEV semantic segmentation map equivalent to a rotation in the x-z plane around the coordinate origin O. 
The rotated GT BEV semantic segmentation map can be obtained by inverse warping operation with a 2D rotation matrix $H_{2}$:
\begin{equation}\label{2d_rotation_matrix}
    H_{2}=\begin{bmatrix}
        cos(\alpha) & -sin(\alpha) \\
        sin(\alpha) & cos(\alpha)
        \end{bmatrix}.
\end{equation}
Note that if the above y-axis is not vertical to the BEV plane, it is impossible to get the rotated BEV segmentation map because of the unpredictable occlusion.

Conjoint rotation acts concurrently on the FV images and the GT BEV segmentation map for labeled data while only acting on the FV images for unlabeled data.

\subsection{Training Process}\label{training_process}
Each training batch consists of half labeled data and half unlabeled data, which is then augmented by the proposed conjoint rotation to get $I_{L}$ and $I_{U}$. And $I^{'}_{U}$ is got by horizontal flipping the $I_{U}$. The Student Net $M_{S}$ is used to predict BEV semantic segmentation maps $Y$ and $Y^{'}_{U}$. The Teacher Net $M_{T}$ is used to predict $Y_{U}$. We update $\theta_{t}$, the parameters of $M_{S}$ at training step $t$ using following overall loss:
\begin{equation}\label{final_loss}
    L=L_{sup}+\lambda_{1}L_{sc}+\lambda_{2}L_{fc}, 
\end{equation}

Then, following Mean-Teacher~\cite{tarvainen2017mean}, we update $\theta^{'}_{t}$, the parameters of $M_{T}$ at training step $t$ using exponential moving average (EMA):
\begin{equation}
    \theta^{'}_{t} = \alpha\theta^{'}_{t-1}+(1-\alpha)\theta_{t}.
\end{equation}
where the EMA decay factor $\alpha$ is set as 0.999 empirically. And after the training, we use the $M_{T}$ for evaluation.

\section{Experiments}

\subsection{Datasets}
We conduct experiments on the nuScenes~\cite{caesar2020nuscenes}.
Following~\cite{PON}, We use the same data generation process and the same data split. 
The training set and testing set contain 168048 images and 35886 images, respectively. 
The resolution of input images is $600\times 800$, and the output BEV semantic segmentation map has a resolution of $196\times 200$, with each pixel representing $0.25m\times0.25m$ in the real world. 

Following the conventions in semi-supervised tasks, we divide the training set into labeled and unlabeled subsets with different ratios. Specifically, we use the first 5\%, 10\%, 20\%, and 40\% samples of each sequence of nuScenes as the labeled set and assume the remaining samples as the unlabeled set. 

\subsection{Network Architecture}
Our model has the same architecture as 
PON~\cite{PON}, a milestone work in the visual BEV semantic segmentation field. The mode uses a ResNet-50 with an FPN~\cite{fpn} as the backbone. The view transformer is implemented by an MLP. And the segmentation decoder consists of a stack of residual blocks and a sigmoid activation function at the last layer.

\begin{table*}[h]
    \begin{center}
    \caption{IoU(\%) on nuScenes with different ratios of labels. "C.V.": construction vehicles, "ped.": pedestrian, "motor": motorcycle.}\label{quantitative_results}
    \resizebox{0.99\linewidth}{!}{
    \begin{tabular}{ccccccccccccccccc}
        \hline
        \multicolumn{1}{c|}{Ratio} & \multicolumn{1}{c|}{Method} & \multicolumn{1}{c|}{Mean} & \multicolumn{1}{c}{drivable}  & \multicolumn{1}{c}{crossing} & \multicolumn{1}{c}{walkway} & \multicolumn{1}{c}{carpark} & \multicolumn{1}{c}{car} & \multicolumn{1}{c}{trunk} & \multicolumn{1}{c}{bus} & \multicolumn{1}{c}{trailer} & \multicolumn{1}{c}{C.V.} & \multicolumn{1}{c}{ped.} & \multicolumn{1}{c}{motor.} & \multicolumn{1}{c}{bike} & \multicolumn{1}{c}{cone} & \multicolumn{1}{c}{barrier}\\
        \hline

        \multicolumn{1}{c|}{\multirow{6}{*}{5\%}} & \multicolumn{1}{c|}{sup-only} & \multicolumn{1}{c|}{14.8} & \multicolumn{1}{c}{57.7} & \multicolumn{1}{c}{25.7} & \multicolumn{1}{c}{30.3} & \multicolumn{1}{c}{24.9} & \multicolumn{1}{c}{29.5} & \multicolumn{1}{c}{9.8} & \multicolumn{1}{c}{5.1} & \multicolumn{1}{c}{4.3} & \multicolumn{1}{c}{0.6} & \multicolumn{1}{c}{2.7} & \multicolumn{1}{c}{0.6} & \multicolumn{1}{c}{1.4} & \multicolumn{1}{c}{6.1} & \multicolumn{1}{c}{8.0}\\
        \multicolumn{1}{c|}{} & \multicolumn{1}{c|}{$\Pi$-Model~\cite{lainetemporal}} & \multicolumn{1}{c|}{15.1} & \multicolumn{1}{c}{57.1} & \multicolumn{1}{c}{26.2} & \multicolumn{1}{c}{29.7} & \multicolumn{1}{c}{24.2} & \multicolumn{1}{c}{29.4} & \multicolumn{1}{c}{10.8} & \multicolumn{1}{c}{5.5} & \multicolumn{1}{c}{7.3} & \multicolumn{1}{c}{\textbf{1.7}} & \multicolumn{1}{c}{2.7} & \multicolumn{1}{c}{0.7} & \multicolumn{1}{c}{1.7} & \multicolumn{1}{c}{5.4} & \multicolumn{1}{c}{8.2}\\
        \multicolumn{1}{c|}{} & \multicolumn{1}{c|}{MT~\cite{tarvainen2017mean}} & \multicolumn{1}{c|}{15.4} & \multicolumn{1}{c}{56.7} & \multicolumn{1}{c}{26.8} & \multicolumn{1}{c}{30.4} & \multicolumn{1}{c}{25.3} & \multicolumn{1}{c}{30.4} & \multicolumn{1}{c}{11.4} & \multicolumn{1}{c}{6.8} & \multicolumn{1}{c}{7.5} & \multicolumn{1}{c}{0.9} & \multicolumn{1}{c}{3.0} & \multicolumn{1}{c}{0.5} & \multicolumn{1}{c}{1.2} & \multicolumn{1}{c}{6.9} & \multicolumn{1}{c}{8.4}\\
        \multicolumn{1}{c|}{} & \multicolumn{1}{c|}{CPS~\cite{chen2021semi}} & \multicolumn{1}{c|}{14.5} & \multicolumn{1}{c}{57.0} & \multicolumn{1}{c}{25.4} & \multicolumn{1}{c}{29.4} & \multicolumn{1}{c}{24.0} & \multicolumn{1}{c}{29.3} & \multicolumn{1}{c}{10.5} & \multicolumn{1}{c}{5.9} & \multicolumn{1}{c}{6.6} & \multicolumn{1}{c}{0.3} & \multicolumn{1}{c}{2.0} & \multicolumn{1}{c}{0.1} & \multicolumn{1}{c}{0.3} & \multicolumn{1}{c}{5.0} & \multicolumn{1}{c}{7.8}\\
        \multicolumn{1}{c|}{} & \multicolumn{1}{c|}{UniMatch~\cite{unimatch}} & \multicolumn{1}{c|}{14.4} & \multicolumn{1}{c}{57.0} & \multicolumn{1}{c}{25.2} & \multicolumn{1}{c}{29.3} & \multicolumn{1}{c}{24.0} & \multicolumn{1}{c}{29.2} & \multicolumn{1}{c}{10.4} & \multicolumn{1}{c}{5.7} & \multicolumn{1}{c}{6.4} & \multicolumn{1}{c}{0.4} & \multicolumn{1}{c}{1.8} & \multicolumn{1}{c}{0.1} & \multicolumn{1}{c}{0.3} & \multicolumn{1}{c}{4.8} & \multicolumn{1}{c}{7.2}\\
        \multicolumn{1}{c|}{} & \multicolumn{1}{c|}{Ours} & \multicolumn{1}{c|}{\textbf{18.1}} & \multicolumn{1}{c}{\textbf{59.3}} & \multicolumn{1}{c}{\textbf{29.8}} & \multicolumn{1}{c}{\textbf{33.8}} & \multicolumn{1}{c}{\textbf{26.1}} & \multicolumn{1}{c}{\textbf{34.6}} & \multicolumn{1}{c}{\textbf{14.7}} & \multicolumn{1}{c}{\textbf{9.8}} & \multicolumn{1}{c}{\textbf{10.7}} & \multicolumn{1}{c}{1.5} & \multicolumn{1}{c}{\textbf{5.9}} & \multicolumn{1}{c}{\textbf{1.6}} & \multicolumn{1}{c}{\textbf{2.7}} & \multicolumn{1}{c}{\textbf{9.6}} & \multicolumn{1}{c}{\textbf{12.5}}\\
        \hline

        \multicolumn{1}{c|}{\multirow{6}{*}{10\%}} & \multicolumn{1}{c|}{sup-only} & \multicolumn{1}{c|}{17.3} & \multicolumn{1}{c}{58.6} & \multicolumn{1}{c}{26.8} & \multicolumn{1}{c}{32.5} & \multicolumn{1}{c}{\textbf{28.5}} & \multicolumn{1}{c}{33.2} & \multicolumn{1}{c}{14.9} & \multicolumn{1}{c}{9.9} & \multicolumn{1}{c}{10.3} & \multicolumn{1}{c}{1.2} & \multicolumn{1}{c}{4.9} & \multicolumn{1}{c}{2.1} & \multicolumn{1}{c}{2.3} & \multicolumn{1}{c}{7.5} & \multicolumn{1}{c}{10.2}\\
        \multicolumn{1}{c|}{} & \multicolumn{1}{c|}{$\Pi$-Model~\cite{lainetemporal}} & \multicolumn{1}{c|}{17.4} & \multicolumn{1}{c}{59.3} & \multicolumn{1}{c}{30.0} & \multicolumn{1}{c}{32.7} & \multicolumn{1}{c}{27.2} & \multicolumn{1}{c}{33.0} & \multicolumn{1}{c}{13.2} & \multicolumn{1}{c}{8.7} & \multicolumn{1}{c}{9.5} & \multicolumn{1}{c}{1.4} & \multicolumn{1}{c}{5.8} & \multicolumn{1}{c}{1.8} & \multicolumn{1}{c}{3.1} & \multicolumn{1}{c}{8.0} & \multicolumn{1}{c}{10.4}\\
        \multicolumn{1}{c|}{} & \multicolumn{1}{c|}{MT~\cite{tarvainen2017mean}} & \multicolumn{1}{c|}{17.8} & \multicolumn{1}{c}{58.6} & \multicolumn{1}{c}{29.7} & \multicolumn{1}{c}{31.8} & \multicolumn{1}{c}{27.4} & \multicolumn{1}{c}{33.7} & \multicolumn{1}{c}{15.6} & \multicolumn{1}{c}{8.9} & \multicolumn{1}{c}{11.4} & \multicolumn{1}{c}{1.9} & \multicolumn{1}{c}{5.1} & \multicolumn{1}{c}{1.8} & \multicolumn{1}{c}{2.8} & \multicolumn{1}{c}{9.4} & \multicolumn{1}{c}{10.8}\\
        \multicolumn{1}{c|}{} & \multicolumn{1}{c|}{CPS~\cite{chen2021semi}} & \multicolumn{1}{c|}{16.4} & \multicolumn{1}{c}{58.7} & \multicolumn{1}{c}{27.0} & \multicolumn{1}{c}{30.3} & \multicolumn{1}{c}{26.0} & \multicolumn{1}{c}{32.0} & \multicolumn{1}{c}{14.6} & \multicolumn{1}{c}{7.7} & \multicolumn{1}{c}{10.1} & \multicolumn{1}{c}{1.2} & \multicolumn{1}{c}{3.0} & \multicolumn{1}{c}{0.5} & \multicolumn{1}{c}{1.3} & \multicolumn{1}{c}{8.0} & \multicolumn{1}{c}{8.7}\\
        \multicolumn{1}{c|}{} & \multicolumn{1}{c|}{UniMatch~\cite{unimatch}} & \multicolumn{1}{c|}{16.6} & \multicolumn{1}{c}{58.8} & \multicolumn{1}{c}{27.2} & \multicolumn{1}{c}{30.2} & \multicolumn{1}{c}{26.4} & \multicolumn{1}{c}{32.4} & \multicolumn{1}{c}{14.7} & \multicolumn{1}{c}{8.0} & \multicolumn{1}{c}{10.0} & \multicolumn{1}{c}{1.7} & \multicolumn{1}{c}{3.2} & \multicolumn{1}{c}{0.7} & \multicolumn{1}{c}{1.2} & \multicolumn{1}{c}{8.5} & \multicolumn{1}{c}{8.9}\\
        \multicolumn{1}{c|}{} & \multicolumn{1}{c|}{Ours}& \multicolumn{1}{c|}{\textbf{20.1}} & \multicolumn{1}{c}{\textbf{60.8}} & \multicolumn{1}{c}{\textbf{31.9}} & \multicolumn{1}{c}{\textbf{35.7}} & \multicolumn{1}{c}{27.4} & \multicolumn{1}{c}{\textbf{36.4}} & \multicolumn{1}{c}{\textbf{17.3}} & \multicolumn{1}{c}{\textbf{13.8}} & \multicolumn{1}{c}{\textbf{13.9}} & \multicolumn{1}{c}{\textbf{2.8}} & \multicolumn{1}{c}{\textbf{7.8}} & \multicolumn{1}{c}{\textbf{4.0}} & \multicolumn{1}{c}{\textbf{5.2}} & \multicolumn{1}{c}{\textbf{11.4}} & \multicolumn{1}{c}{\textbf{12.9}}\\
        \hline

        \multicolumn{1}{c|}{\multirow{6}{*}{20\%}} & \multicolumn{1}{c|}{sup-only} & \multicolumn{1}{c|}{19.5} & \multicolumn{1}{c}{61.0} & \multicolumn{1}{c}{32.4} & \multicolumn{1}{c}{34.8} & \multicolumn{1}{c}{27.7} & \multicolumn{1}{c}{36.8} & \multicolumn{1}{c}{15.8} & \multicolumn{1}{c}{14.0} & \multicolumn{1}{c}{11.9} & \multicolumn{1}{c}{2.3} & \multicolumn{1}{c}{7.1} & \multicolumn{1}{c}{3.8} & \multicolumn{1}{c}{3.5} & \multicolumn{1}{c}{7.8} & \multicolumn{1}{c}{13.7}\\
        \multicolumn{1}{c|}{} & \multicolumn{1}{c|}{$\Pi$-Model~\cite{lainetemporal}} & \multicolumn{1}{c|}{19.8} & \multicolumn{1}{c}{60.7} & \multicolumn{1}{c}{32.2} & \multicolumn{1}{c}{35.2} & \multicolumn{1}{c}{26.8} & \multicolumn{1}{c}{36.5} & \multicolumn{1}{c}{16.7} & \multicolumn{1}{c}{13.6} & \multicolumn{1}{c}{11.4} & \multicolumn{1}{c}{1.9} & \multicolumn{1}{c}{6.8} & \multicolumn{1}{c}{4.7} & \multicolumn{1}{c}{4.7} & \multicolumn{1}{c}{10.6} & \multicolumn{1}{c}{15.0}\\
        \multicolumn{1}{c|}{} & \multicolumn{1}{c|}{MT~\cite{tarvainen2017mean}} & \multicolumn{1}{c|}{20.3} & \multicolumn{1}{c}{60.4} & \multicolumn{1}{c}{32.8} & \multicolumn{1}{c}{36.0} & \multicolumn{1}{c}{29.8} & \multicolumn{1}{c}{36.2} & \multicolumn{1}{c}{17.6} & \multicolumn{1}{c}{11.9} & \multicolumn{1}{c}{12.9} & \multicolumn{1}{c}{\textbf{4.4}} & \multicolumn{1}{c}{7.8} & \multicolumn{1}{c}{3.7} & \multicolumn{1}{c}{4.4} & \multicolumn{1}{c}{\textbf{11.1}} & \multicolumn{1}{c}{15.1}\\
        \multicolumn{1}{c|}{} & \multicolumn{1}{c|}{CPS~\cite{chen2021semi}} & \multicolumn{1}{c|}{18.4} & \multicolumn{1}{c}{59.4} & \multicolumn{1}{c}{31.7} & \multicolumn{1}{c}{35.2} & \multicolumn{1}{c}{29.2} & \multicolumn{1}{c}{35.0} & \multicolumn{1}{c}{17.0} & \multicolumn{1}{c}{10.9} & \multicolumn{1}{c}{12.0} & \multicolumn{1}{c}{3.6} & \multicolumn{1}{c}{7.0} & \multicolumn{1}{c}{2.7} & \multicolumn{1}{c}{3.5} & \multicolumn{1}{c}{10.2} & \multicolumn{1}{c}{13.6}\\
        \multicolumn{1}{c|}{} & \multicolumn{1}{c|}{UniMatch~\cite{unimatch}} & \multicolumn{1}{c|}{18.2} & \multicolumn{1}{c}{59.4} & \multicolumn{1}{c}{31.3} & \multicolumn{1}{c}{35.0} & \multicolumn{1}{c}{29.0} & \multicolumn{1}{c}{35.0} & \multicolumn{1}{c}{16.8} & \multicolumn{1}{c}{10.5} & \multicolumn{1}{c}{11.8} & \multicolumn{1}{c}{3.5} & \multicolumn{1}{c}{6.8} & \multicolumn{1}{c}{2.6} & \multicolumn{1}{c}{3.2} & \multicolumn{1}{c}{10.0} & \multicolumn{1}{c}{13.6}\\
        \multicolumn{1}{c|}{} & \multicolumn{1}{c|}{Ours} & \multicolumn{1}{c|}{\textbf{21.9}} & \multicolumn{1}{c}{\textbf{61.5}} & \multicolumn{1}{c}{\textbf{34.0}} & \multicolumn{1}{c}{\textbf{37.0}} & \multicolumn{1}{c}{\textbf{30.6}} & \multicolumn{1}{c}{\textbf{38.5}} & \multicolumn{1}{c}{\textbf{20.4}} & \multicolumn{1}{c}{\textbf{16.8}} & \multicolumn{1}{c}{\textbf{14.3}} & \multicolumn{1}{c}{3.4} & \multicolumn{1}{c}{\textbf{9.7}} & \multicolumn{1}{c}{\textbf{6.7}} & \multicolumn{1}{c}{\textbf{6.9}} & \multicolumn{1}{c}{10.7} & \multicolumn{1}{c}{\textbf{15.8}}\\
        \hline

        \multicolumn{1}{c|}{\multirow{6}{*}{40\%}} & \multicolumn{1}{c|}{sup-only} & \multicolumn{1}{c|}{22.3} & \multicolumn{1}{c}{61.3} & \multicolumn{1}{c}{34.9} & \multicolumn{1}{c}{37.5} & \multicolumn{1}{c}{30.9} & \multicolumn{1}{c}{38.9} & \multicolumn{1}{c}{20.5} & \multicolumn{1}{c}{17.8} & \multicolumn{1}{c}{16.4} & \multicolumn{1}{c}{3.0} & \multicolumn{1}{c}{10.5} & \multicolumn{1}{c}{6.1} & \multicolumn{1}{c}{6.3} & \multicolumn{1}{c}{\textbf{11.5}} & \multicolumn{1}{c}{15.9}\\
        \multicolumn{1}{c|}{} & \multicolumn{1}{c|}{$\Pi$-Model~\cite{lainetemporal}} & \multicolumn{1}{c|}{22.6} & \multicolumn{1}{c}{61.8} & \multicolumn{1}{c}{35.1} & \multicolumn{1}{c}{37.9} & \multicolumn{1}{c}{30.8} & \multicolumn{1}{c}{38.2} & \multicolumn{1}{c}{20.6} & \multicolumn{1}{c}{21.1} & \multicolumn{1}{c}{15.5} & \multicolumn{1}{c}{4.7} & \multicolumn{1}{c}{9.9} & \multicolumn{1}{c}{6.4} & \multicolumn{1}{c}{7.1} & \multicolumn{1}{c}{10.4} & \multicolumn{1}{c}{16.4}\\
        \multicolumn{1}{c|}{} & \multicolumn{1}{c|}{MT~\cite{tarvainen2017mean}} & \multicolumn{1}{c|}{22.6} & \multicolumn{1}{c}{61.5} & \multicolumn{1}{c}{34.9} & \multicolumn{1}{c}{37.9} & \multicolumn{1}{c}{\textbf{31.6}} & \multicolumn{1}{c}{38.4} & \multicolumn{1}{c}{20.0} & \multicolumn{1}{c}{18.5} & \multicolumn{1}{c}{16.2} & \multicolumn{1}{c}{3.2} & \multicolumn{1}{c}{10.5} & \multicolumn{1}{c}{\textbf{7.6}} & \multicolumn{1}{c}{\textbf{8.3}} & \multicolumn{1}{c}{10.9} & \multicolumn{1}{c}{\textbf{16.7}}\\
        \multicolumn{1}{c|}{} & \multicolumn{1}{c|}{CPS~\cite{chen2021semi}} & \multicolumn{1}{c|}{20.6} & \multicolumn{1}{c}{59.5} & \multicolumn{1}{c}{33.0} & \multicolumn{1}{c}{36.0} & \multicolumn{1}{c}{29.5} & \multicolumn{1}{c}{37.3} & \multicolumn{1}{c}{18.0} & \multicolumn{1}{c}{17.2} & \multicolumn{1}{c}{13.1} & \multicolumn{1}{c}{1.6} & \multicolumn{1}{c}{8.5} & \multicolumn{1}{c}{5.7} & \multicolumn{1}{c}{6.2} & \multicolumn{1}{c}{8.9} & \multicolumn{1}{c}{15.0}\\
        \multicolumn{1}{c|}{} & \multicolumn{1}{c|}{UniMatch~\cite{unimatch}} & \multicolumn{1}{c|}{20.5} & \multicolumn{1}{c}{59.6} & \multicolumn{1}{c}{33.0} & \multicolumn{1}{c}{35.7} & \multicolumn{1}{c}{29.4} & \multicolumn{1}{c}{37.3} & \multicolumn{1}{c}{18.0} & \multicolumn{1}{c}{17.0} & \multicolumn{1}{c}{12.8} & \multicolumn{1}{c}{1.6} & \multicolumn{1}{c}{8.4} & \multicolumn{1}{c}{5.6} & \multicolumn{1}{c}{6.0} & \multicolumn{1}{c}{8.9} & \multicolumn{1}{c}{15.0}\\
        \multicolumn{1}{c|}{} & \multicolumn{1}{c|}{Ours} & \multicolumn{1}{c|}{\textbf{23.5}} & \multicolumn{1}{c}{\textbf{62.8}} & \multicolumn{1}{c}{\textbf{36.0}} & \multicolumn{1}{c}{\textbf{38.9}} & \multicolumn{1}{c}{31.5} & \multicolumn{1}{c}{\textbf{39.6}} & \multicolumn{1}{c}{\textbf{22.7}} & \multicolumn{1}{c}{\textbf{21.6}} & \multicolumn{1}{c}{\textbf{18.3}} & \multicolumn{1}{c}{\textbf{5.1}} & \multicolumn{1}{c}{\textbf{11.1}} & \multicolumn{1}{c}{6.8} & \multicolumn{1}{c}{7.6} & \multicolumn{1}{c}{11.1} & \multicolumn{1}{c}{16.1}\\
        \hline
    \end{tabular}}
    
    \end{center}
    \vspace{-0.4cm}
\end{table*}

\begin{figure*}[h]
    \begin{center}
    \includegraphics[width=17.5cm]{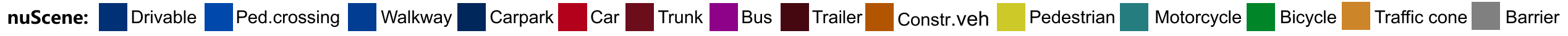}\\
    \subfigure{
        \begin{minipage}[t]{0.133\linewidth}
            \centering
            \includegraphics[width=2.7cm]{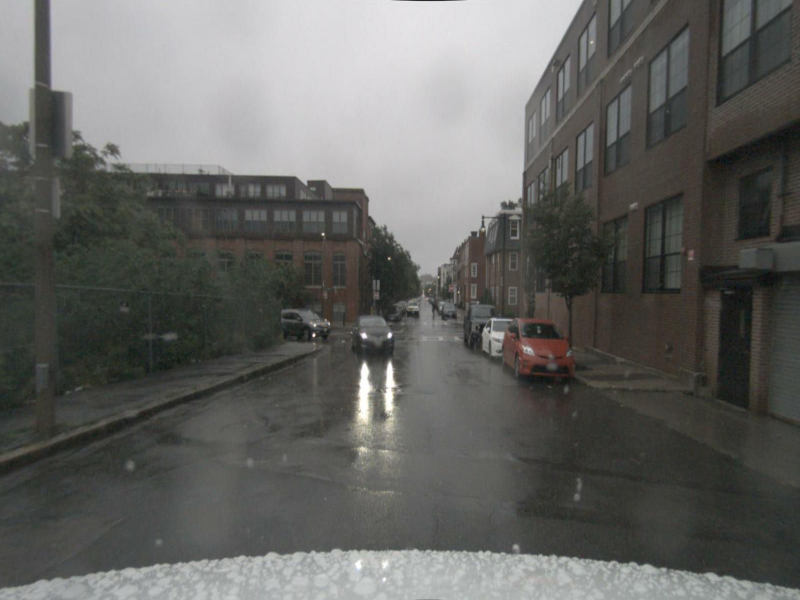}\\
            \vspace{0.09cm}
            \scriptsize{FV image}
        \end{minipage}
    } 
    \subfigure{
        \begin{minipage}[t]{0.10\linewidth}
            \centering
            \includegraphics[width=2.1cm]{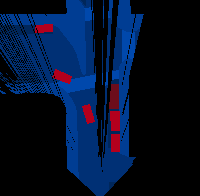}\\
            \vspace{0.06cm}
            \scriptsize{GT}
        \end{minipage}
    } 
    \subfigure{
        \begin{minipage}[t]{0.10\linewidth}
            \centering
            \includegraphics[width=2.1cm]{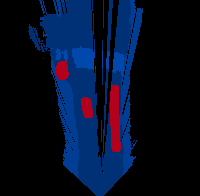}\\
            \vspace{0.06cm}
            \scriptsize{sup-only}
        \end{minipage}
    }
    \subfigure{
        \begin{minipage}[t]{0.10\linewidth}
            \centering
            \includegraphics[width=2.1cm]{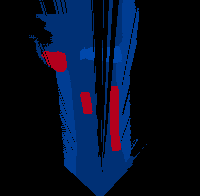}\\
            \vspace{0.06cm}
            \scriptsize{$\Pi$-Model\cite{lainetemporal}}
        \end{minipage}
    } 
    \subfigure{
        \begin{minipage}[t]{0.10\linewidth}
            \centering
            \includegraphics[width=2.1cm]{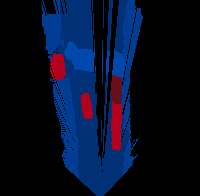}\\
            \vspace{0.06cm}
            \scriptsize{MT\cite{tarvainen2017mean}}
        \end{minipage}
    }
    \subfigure{
        \begin{minipage}[t]{0.10\linewidth}
            \centering
            \includegraphics[width=2.1cm]{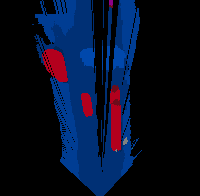}\\
            \vspace{0.06cm}
            \scriptsize{CPS\cite{chen2021semi}}
        \end{minipage}
    }
    \subfigure{
        \begin{minipage}[t]{0.10\linewidth}
            \centering
            \includegraphics[width=2.1cm]{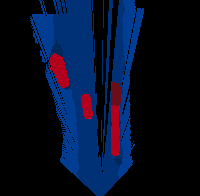}\\
            \vspace{0.06cm}
            \scriptsize{UniMatch\cite{unimatch}}
        \end{minipage}
    }
    \subfigure{
        \begin{minipage}[t]{0.10\linewidth}
            \centering
            \includegraphics[width=2.1cm]{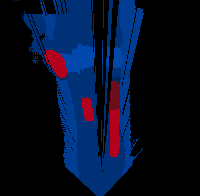}\\
            \vspace{0.06cm}
            \scriptsize{Ours}
        \end{minipage}
    }
    \end{center}
    \vspace{-0.4cm}
    \caption{\textbf{Qualitative results with 20\% labels.} We follow the color scheme in PON~\cite{PON} and use the visibility mask (black) for visualization.}\label{qualitative_results}
    \vspace{-0.6cm}
\end{figure*}

\subsection{Implementation Details}
Our work is implemented in Pytorch on 8 NVIDIA V100 GPUs. We train the models using the Adam optimizer with 25 epochs and a batch size of 32. 
The initial learning rate is set as $1\times 10^{-4}$ and decays to $1\times 10^{-5}$ after 15 epochs. 
The weight $\lambda_{1}$ and $\lambda_{2}$ in Eq.~\ref{final_loss} are empirically set as $2\times 10^{-3}$ and $2\times 10^{-4}$ respectively. 
Besides, we set $\alpha_{max}=35^{\circ}$ for the proposed conjoint rotation augmentation. With a 50\% chance, we apply the conjoint rotation on the FV images before being resized and fed to the network. For evaluation, following~\cite{PON}, we use the IoU as our evaluation metric, and those invisible pixels are ignored during evaluation.

For sufficient comparison, we extend several classical and state-of-the-art semi-supervised 2D segmentation methods to this task. Especially, we look into $\Pi$-Model~\cite{lainetemporal}, Mean-Teacher (MT)~\cite{tarvainen2017mean}, CPS~\cite{chen2021semi} and UniMatch~\cite{unimatch}. And their weights of consistency loss are respectively set as $5\times 10^{-3}$, $2\times 10^{-3}$, $1\times 10^{-3}$ and $1\times 10^{-3}$.

\subsection{Main Results}
Tab.~\ref{quantitative_results} presents the class-wise IoU scores on the nuScenes dataset. With various ratios of labeled data, our semi-supervised framework can significantly outperform the supervised-only baselines in almost all categories, indicating that our framework is able to utilize unlabeled data to effectively enhance performance. Given 10\% labeled data and 90\% unlabeled data, our framework can even outperform the full-supervised baseline using 20\% labels. A similar improvement is also achieved under the setting of 5\% labels. This implies that our semi-supervised framework can enhance the efficiency of data utilization. Compared with extended 2D semi-supervised methods~\cite{lainetemporal,tarvainen2017mean,chen2021semi,unimatch}, our framework achieves better scores. Thus the superiority of our framework is presented. And interestingly, state-of-the-art 2D semi-supervised methods~\cite{chen2021semi,unimatch} perform worse than classical methods~\cite{lainetemporal,tarvainen2017mean} on this task. We conjecture that the operation of generating pseudo labels in \cite{chen2021semi} and \cite{unimatch} is not applicable to this multi-label classification task.

Furthermore, the qualitative results on the nuScenes with 20\% labels are shown in Fig.~\ref{qualitative_results}. They also prove that by exploiting unlabeled data, our framework helps improve the semantic segmentation quality.

\begin{table}[h]
    \vspace{-0.2cm}
    \caption{mIoU(\%) for ablation studies. CR denotes conjoint rotation.}\label{component_ablation}
    \vspace{-0.6cm}
    \begin{center}
    \resizebox{1.0\linewidth}{!}{
    \begin{tabular}{cccccccc}
        \toprule
        \multicolumn{1}{c}{$L_{sup}$} & \multicolumn{1}{c}{$L_{sc}$} & \multicolumn{1}{c}{$L_{fc}$} & \multicolumn{1}{c|}{CR} & \multicolumn{1}{c}{5\%} & \multicolumn{1}{c}{10\%}\ & \multicolumn{1}{c}{20\%} & \multicolumn{1}{c}{40\%}\\
        \midrule
        \multicolumn{1}{c}{\checkmark} & \multicolumn{1}{c}{} & \multicolumn{1}{c}{} & \multicolumn{1}{c|}{} & \multicolumn{1}{c}{14.8} & \multicolumn{1}{c}{17.3} & \multicolumn{1}{c}{19.5} & \multicolumn{1}{c}{22.3}\\
        \multicolumn{1}{c}{\checkmark} & \multicolumn{1}{c}{\checkmark} & \multicolumn{1}{c}{} & \multicolumn{1}{c|}{} & \multicolumn{1}{c}{15.4\textcolor{blue}{($\uparrow$0.6)}} & \multicolumn{1}{c}{17.8\textcolor{blue}{($\uparrow$0.5)}} & \multicolumn{1}{c}{20.3\textcolor{blue}{($\uparrow$0.8)}} & \multicolumn{1}{c}{22.6\textcolor{blue}{($\uparrow$0.3)}}\\
        \multicolumn{1}{c}{\checkmark} & \multicolumn{1}{c}{\checkmark} & \multicolumn{1}{c}{\checkmark} & \multicolumn{1}{c|}{} & \multicolumn{1}{c}{15.5\textcolor{blue}{($\uparrow$0.7)}} & \multicolumn{1}{c}{17.9\textcolor{blue}{($\uparrow$0.6)}} & \multicolumn{1}{c}{20.5\textcolor{blue}{($\uparrow$1.0)}} & \multicolumn{1}{c}{22.3\textcolor{blue}{($\uparrow$0.0)}}\\
        \multicolumn{1}{c}{\checkmark} & \multicolumn{1}{c}{} & \multicolumn{1}{c}{} & \multicolumn{1}{c|}{\checkmark} & \multicolumn{1}{c}{17.5\textcolor{blue}{($\uparrow$2.7)}} & \multicolumn{1}{c}{19.2\textcolor{blue}{($\uparrow$1.9)}} & \multicolumn{1}{c}{21.1\textcolor{blue}{($\uparrow$1.6)}} & \multicolumn{1}{c}{22.8\textcolor{blue}{($\uparrow$0.5)}}\\
        \multicolumn{1}{c}{\checkmark} & \multicolumn{1}{c}{\checkmark} & \multicolumn{1}{c}{} & \multicolumn{1}{c|}{\checkmark} & \multicolumn{1}{c}{18.1\textcolor{blue}{($\uparrow$3.3)}} & \multicolumn{1}{c}{20.1\textcolor{blue}{($\uparrow$2.8)}} & \multicolumn{1}{c}{21.4\textcolor{blue}{($\uparrow$1.9)}} & \multicolumn{1}{c}{23.3\textcolor{blue}{($\uparrow$1.0)}}\\
        \multicolumn{1}{c}{\checkmark} & \multicolumn{1}{c}{\checkmark} & \multicolumn{1}{c}{\checkmark} & \multicolumn{1}{c|}{\checkmark} & \multicolumn{1}{c}{\textbf{18.1\textcolor{blue}{($\uparrow$3.3)}}} & \multicolumn{1}{c}{\textbf{20.1\textcolor{blue}{($\uparrow$2.8)}}} & \multicolumn{1}{c}{\textbf{21.9\textcolor{blue}{($\uparrow$2.4)}}} & \multicolumn{1}{c}{\textbf{23.5\textcolor{blue}{($\uparrow$1.2)}}}\\
        \bottomrule
    \end{tabular}}
    \end{center}
    \vspace{-0.6cm}
\end{table}

\subsection{Ablation Study}
To better understand the effect of each component of our framework, we conduct an ablation study as presented in Tab.~\ref{component_ablation}. The results show that when all components are combined together, the performance is the best. Note that we only report mIoU scores in this section, due to the abundance of experiments.

\begin{figure*}[h]
    \begin{center}
    \subfigure{
        \begin{minipage}[t]{0.18\linewidth}
        \centering
        \begin{overpic}[width=3.0cm]{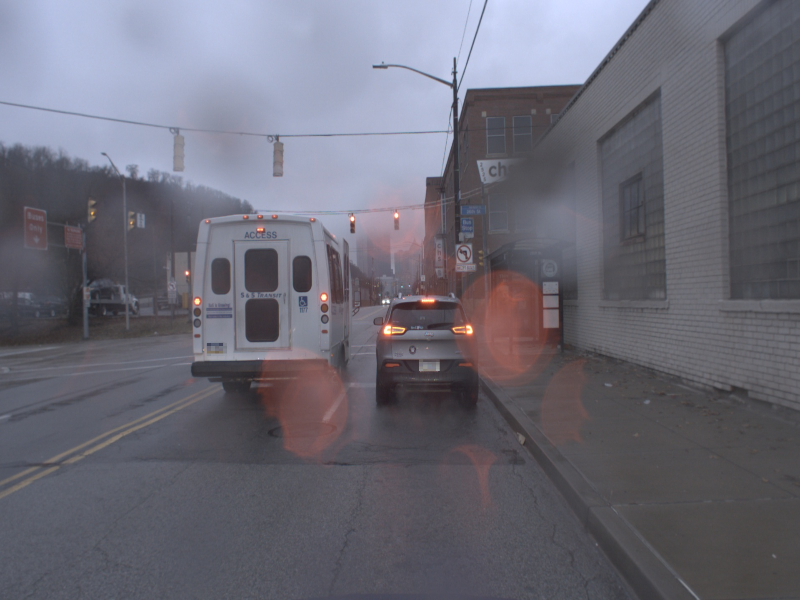}
            \put(3, 65){\textcolor{red}{(a)}}
        \end{overpic}\\
        \end{minipage}
    }
    \subfigure{
        \begin{minipage}[t]{0.18\linewidth}
        \centering
        \begin{overpic}[width=3.0cm]{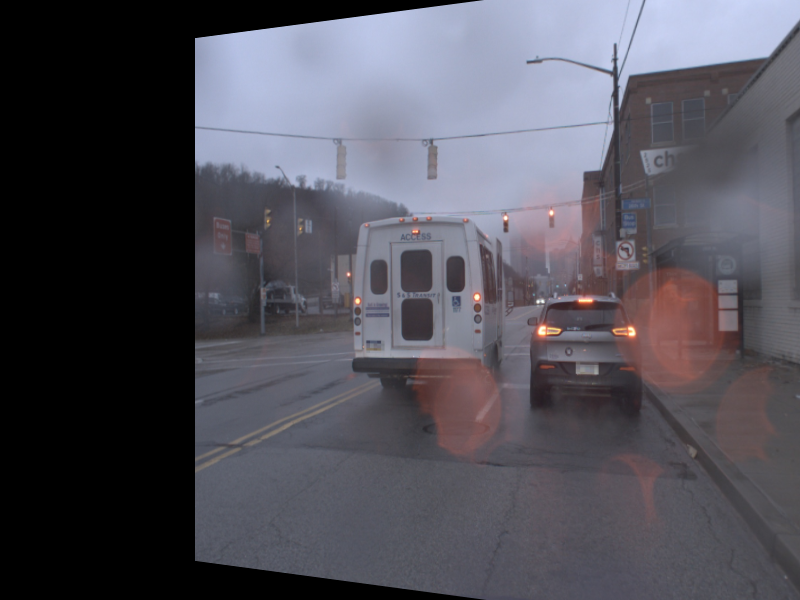}
            \put(3, 65){\textcolor{red}{(b)}}
        \end{overpic}\\
        \end{minipage}
    }
    \subfigure{
        \begin{minipage}[t]{0.18\linewidth}
        \centering
        \begin{overpic}[width=3.0cm]{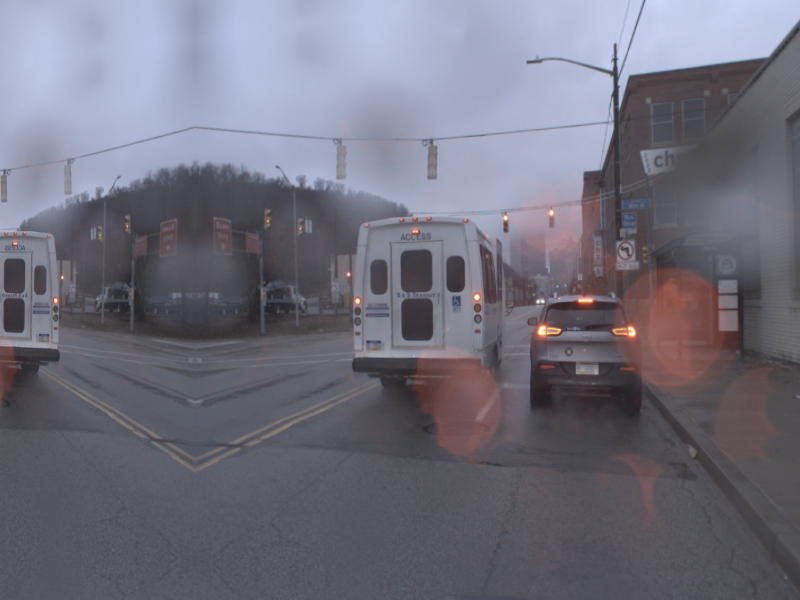}
            \put(3, 65){\textcolor{red}{(c)}}
        \end{overpic}\\
        \end{minipage}
    }
    \subfigure{
        \begin{minipage}[t]{0.18\linewidth}
        \centering
        \begin{overpic}[width=3.0cm]{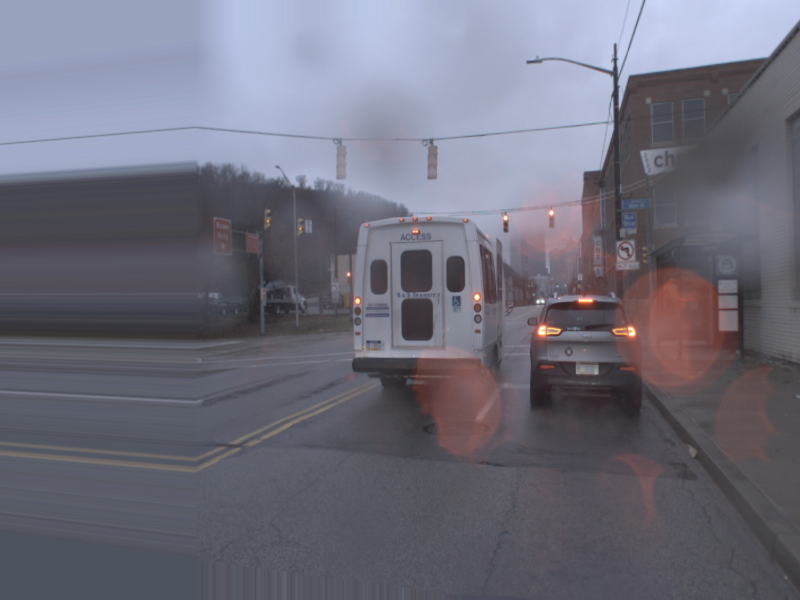}
            \put(3, 65){\textcolor{red}{(d)}}
        \end{overpic}\\
        \end{minipage}
    }
    \end{center}
    \vspace{-0.4cm}
    \caption{\textbf{Different border modes.} (a)Original FV image. Augmented FV image with (b)zero border, (c)reflect border, and (d)replicate border.}\label{border_mode_vizs}
    \vspace{-0.6cm}
\end{figure*}

We conduct more detailed ablation studies to get deeper insights into our framework. The following experiments are conducted with 20\% labels unless otherwise specified.

\noindent\textbf{Benefits of Consistency losses.} The supervised loss $L_{sup}$ gives the model the primary supervisory signal. However, when labels are limited, information of unlabeled data cannot be excavated with only $L_{sup}$. 
As shown in Tab.~\ref{component_ablation}, compared with the full-supervised model, better scores can be gained by introducing segmentation consistency loss $L_{sc}$. Furthermore, applying consistency constraints on the BEV feature by $L_{fc}$, model performance can be further refined in almost all cases. 

\begin{figure}[h]
    \begin{center}
    \includegraphics[scale=0.25]{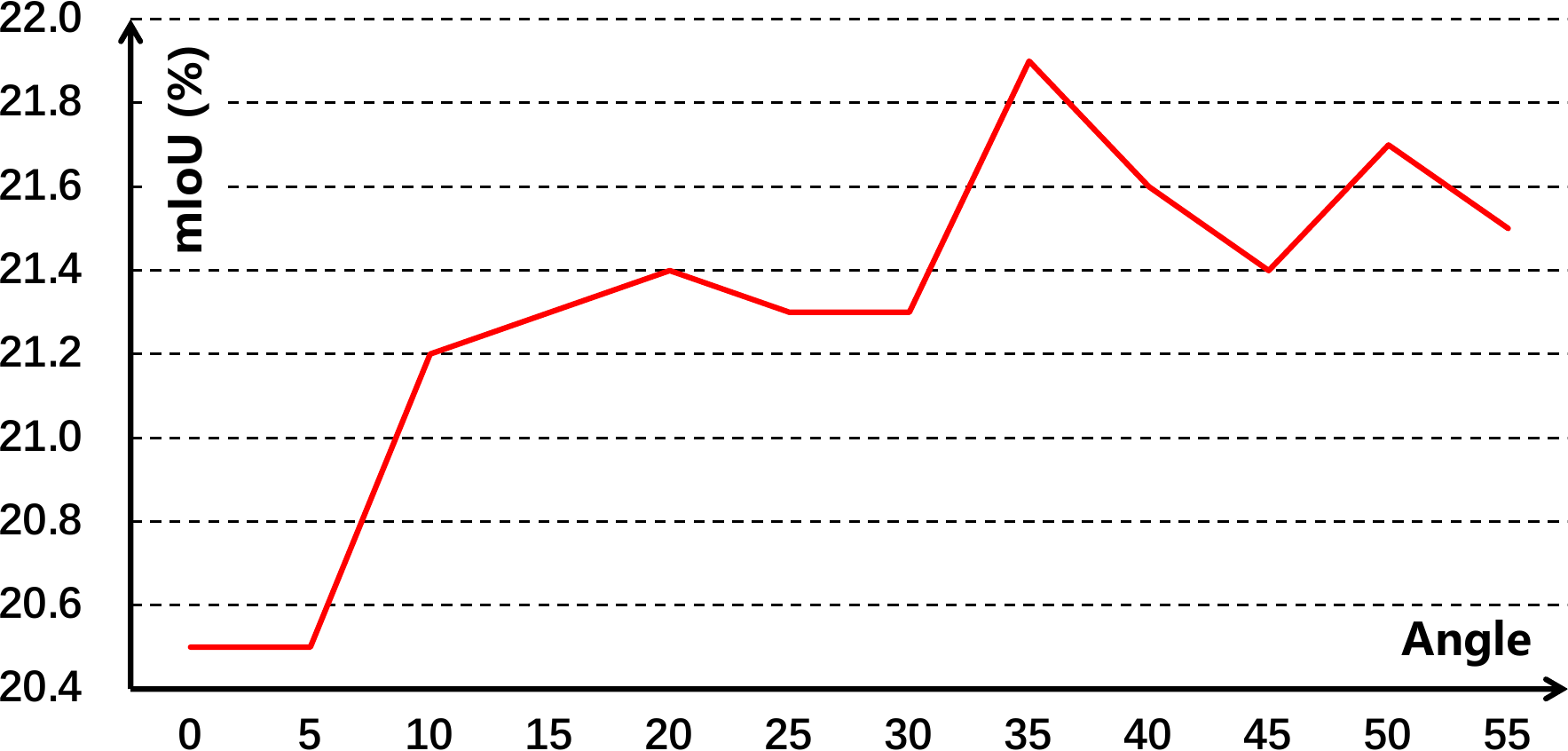}
    \end{center}
    \vspace{-0.4cm}
    \caption{\textbf{Sensitivity of $\theta_{max}$ of conjoint rotation augmentation.}}\label{rotation_angle_ablation}
    \vspace{-0.4cm}
\end{figure}

\noindent\textbf{Effectiveness of conjoint rotation.}\label{conjoint_rotation_ablation} The results in Tab.~\ref{component_ablation} show that in all cases, conjoint rotation can significantly improve performance. Thus, the conjoint rotation is effective for both full-supervised and semi-supervised models thanks to maintaining the 3D geometric relationship even though information on image edges may be lost. As the unique hyper-parameter for conjoint rotation, the role of $\alpha_{max}$ needs to be explored. And we conduct sensitivity experiments on $\alpha_{max}$ and show results in Fig.~\ref{rotation_angle_ablation}. According to Fig.~\ref{rotation_angle_ablation}, we choose $35^\circ$ as $\theta_{max}$ for better performance. And the results also show that the improvement is remarkable in a wide range of $\theta_{max}$ ($15^{\circ}-55^{\circ}$), which validates the robustness of conjoint rotation. The bordering mode is also essential for conjoint rotation. We compare the performance using different bordering modes(Fig.~\ref{border_mode_vizs}(b) to (d)) and present the results in Tab.~\ref{border_mode_ablation}. With zero border, image black edges brought by forward warping operation can lower the improvement of conjoint rotation. And replicating border can make the improvement more significant. But interestingly, there is no improvement when reflecting border. Moreover, we make a comparison with other augmentation methods to demonstrate our effectiveness in Tab.~\ref{augmentation_ablation}. The unsatisfaction with Cutout~\cite{devries2017improved} and Random Erasing~\cite{zhong2020random} may lie in the damage to the geometric relationship between the FV images and the BEV semantic segmentation maps. Although BEV-Space data augmentation~\cite{huang2021bevdet} can improve detection performance, the performance degradation shown in Tab.~\ref{augmentation_ablation} proves that it's not applicable to this task.

\begin{table}[h]
    \vspace{-0.2cm}
    \caption{\textbf{Abation study on border mode.}}\label{border_mode_ablation}
    \vspace{-0.4cm}
    \begin{center}
    \resizebox{0.85\linewidth}{!}{
    \begin{tabular}{ccc}
        \toprule
        \multicolumn{1}{c}{Replicate Border (Ours)} & \multicolumn{1}{c}{Zero Border} & \multicolumn{1}{c}{Reflect Border} \\
        \midrule
        \multicolumn{1}{c}{\textbf{21.9}} & \multicolumn{1}{c}{21.3} & \multicolumn{1}{c}{21.3}\\
        \bottomrule
    \end{tabular}}
    \end{center}
    \vspace{-0.5cm}
\end{table}

\begin{table}[h]
    \caption{\textbf{Comparison with other augmentation methods.}}\label{augmentation_ablation}
    \vspace{-0.4cm}
    \begin{center}
    \resizebox{0.85\linewidth}{!}{
    \begin{tabular}{cc}
        \toprule
        \multicolumn{1}{c|}{Augmentation method} & \multicolumn{1}{c}{mIoU(\%)} \\
        \midrule
        \multicolumn{1}{c|}{Cutout~\cite{devries2017improved}} & \multicolumn{1}{c}{20.1} \\
        \multicolumn{1}{c|}{Random Erasing~\cite{zhong2020random}} & \multicolumn{1}{c}{20.5} \\
        \multicolumn{1}{c|}{BEV-space Data Augmentation(Rotate)~\cite{huang2021bevdet}} & \multicolumn{1}{c}{20.4} \\
        \multicolumn{1}{c|}{BEV-space Data Augmentation(Flip)~\cite{huang2021bevdet}} & \multicolumn{1}{c}{20.4} \\
        \multicolumn{1}{c|}{Conjoint Rotation} & \multicolumn{1}{c}{\textbf{21.9}} \\
        \bottomrule
    \end{tabular}}
    \end{center}
    \vspace{-0.8cm}
\end{table}

\noindent\textbf{Perturbation strategy.} Different perturbation strategies may bring different results for consistency-based semi-supervised learning. We make a comparison between our horizontal flip and color jitter, a common perturbation strategy in the semi-supervised 2D semantic segmentation field. Results of the first two rows in Tab.~\ref{perturbation_ablation} show that our framework without $L_{fc}$ performs slightly better when using color jitter as the perturbation. But when feature consistency loss $L_{fc}$ is applied, performance can be further improved with the horizontal flip while almost unchanged with color jitter. It indicates that $L_{fc}$ improves the performance by effectively constraining the spatial consistency that is perturbed by horizontal flip. And the results in the third row imply that the color jitter can destroy such consistency.

\begin{table}[h]
    \vspace{-0.2cm}
    \caption{\textbf{mIoU(\%) scores with different perturbation strategy.}}\label{perturbation_ablation}
    \vspace{-0.4cm}
    \begin{center}
    \resizebox{0.8\linewidth}{!}{
    \begin{tabular}{cccc}
        \toprule
        \multicolumn{1}{c}{Horizontal Flip} & \multicolumn{1}{c|}{Color Jitter} & \multicolumn{1}{c}{Ours w/o $L_{fc}$} & \multicolumn{1}{c}{Ours}\\
        \midrule
        \multicolumn{1}{c}{\checkmark} & \multicolumn{1}{c|}{} & \multicolumn{1}{c}{21.4} & \multicolumn{1}{c}{\textbf{21.9}} \\
        \multicolumn{1}{c}{} & \multicolumn{1}{c|}{\checkmark} & \multicolumn{1}{c}{21.5} & \multicolumn{1}{c}{21.5} \\
        \multicolumn{1}{c}{\checkmark} & \multicolumn{1}{c|}{\checkmark} & \multicolumn{1}{c}{21.3} & \multicolumn{1}{c}{21.6} \\
        \bottomrule
    \end{tabular}}
    \end{center}
    \vspace{-0.5cm}
\end{table}

\noindent\textbf{Improvements with 3D-to-2D-based VT.} To verify the effectiveness of our framework, we further use the 3D-to-2D-based VT~\cite{li2022bevformer} to conduct the experiments. mIoU scores in Tab.~\ref{improvements_on_other_VTS} validate that our framework can still effectively exploit unlabeled data to improve performance. 

\begin{table}[h]
    \vspace{-0.2cm}
    \caption{Improvements with 3D-to-2D-based VT on the nuScenes}\label{improvements_on_other_VTS}
    \vspace{-0.4cm}
    \begin{center}
    \resizebox{0.85\linewidth}{!}{
    \begin{tabular}{ccccc}
        \toprule
        \multicolumn{1}{c|}{Method} & \multicolumn{1}{c}{5\%} & \multicolumn{1}{c}{10\%}\ & \multicolumn{1}{c}{20\%} & \multicolumn{1}{c}{40\%}\\
        \midrule
        \multicolumn{1}{c|}{3D-to-2D(sup-only)} & \multicolumn{1}{c}{14.6} & \multicolumn{1}{c}{16.4} & \multicolumn{1}{c}{17.6} & \multicolumn{1}{c}{19.9}\\
        \multicolumn{1}{c|}{3D-to-2D(semi-sup)} & \multicolumn{1}{c}{\textbf{16.2}} & \multicolumn{1}{c}{\textbf{17.4}} & \multicolumn{1}{c}{\textbf{18.9}} & \multicolumn{1}{c}{\textbf{20.7}}\\
        \bottomrule
    \end{tabular}}
    \end{center}
    \vspace{-0.6cm}
\end{table}

\section{Conclusion}
In this work, we delve into the visual BEV semantic segmentation with limited labels and present a novel semi-supervised framework to utilize unlabeled data to improve performance. We propose restricting the model using consistency on semantic segmentation and the BEV feature to use unlabeled data fully. Moreover, we design a novel data augmentation method based on the ingenious geometric relationship. Experiment results demonstrate that our framework can effectively improve performance and data utilization even when using different view transformer. And the effectiveness of our contributions is proved by extensive ablation studies. 
In the future, we will investigate extending our contributions to BEV detection and 3D semantic occupancy tasks. 

{
\small
\bibliographystyle{ieeetr}
\bibliography{sections/reference}
}

\end{document}